\documentclass[11pt]{article}

\usepackage[final]{acl}

\usepackage{times}
\usepackage{latexsym}

\usepackage[T1]{fontenc}

\usepackage[utf8]{inputenc}

\usepackage{microtype}

\usepackage{inconsolata}

\usepackage{graphicx}
\usepackage{multirow}
\usepackage{booktabs}
\usepackage{enumitem}
\usepackage{amsmath}
\usepackage{xspace}
\usepackage{amssymb}
\usepackage[ruled,vlined]{algorithm2e}
\usepackage{setspace}
\usepackage{bm}
\usepackage{soul}      
\usepackage{xcolor}    
\usepackage{framed}    
\colorlet{shadecolor}{yellow!30}  
\sethlcolor{yellow}     
\newcommand{\ourtitle}{EKSFT}
\title{Entropy-KL Divergence-based Token Masking: A Novel Approach for Selective Fine-tuning of Large Language Models}

\author{
    \textbf{Qi Liu}\textsuperscript{1},
    \textbf{Mingdi Sun}\textsuperscript{1},
    \textbf{Yongyi He}\textsuperscript{1},
    \textbf{Zhi Zheng}\textsuperscript{1},
    \textbf{Tong Xu}\textsuperscript{1},
    \textbf{Yi Zheng}\textsuperscript{2}, \\
    \textbf{Zhefeng Wang}\textsuperscript{2},
    \textbf{Enhong Chen}\textsuperscript{1} \\
    \textsuperscript{1} University of Science and Technology of China \\
    \textsuperscript{2} Huawei Cloud \\
    \texttt{\{liuqilq, sun-123, vagabond\}@mail.ustc.edu.cn} \\
    \texttt{\{zhengzhi97, tongxu, cheneh\}@ustc.edu.cn} \\
    \texttt{\{zhengyi29, wangzhefeng\}@huawei.com} 
}

\begin{document}
\maketitle

\begin{abstract}
Supervised fine-tuning (SFT) followed by reinforcement learning (RL)
has become a standard post-training paradigm for large language models.
This paradigm provides a cold-start for RL exploration,
avoiding the inefficiency of pure RL where on-policy sampling yields insufficient positive samples.
However, in practice, existing approaches often use a small amount of data for 
SFT initialization compared to the RL phase,
which can cause the model to fit the limited samples 
and shift away from its pre-trained distribution.
This distribution shift impedes the model's ability to effectively explore during subsequent RL training.
To address this challenge, we propose that in low-data regimes,
SFT should prioritize \emph{activating} task-relevant capabilities
rather than \emph{memorizing} specific content.
Along this line, we propose \textbf{\ourtitle} (Entropy-KL Selective Fine-Tuning),
which selectively masks tokens that exhibit either high entropy
or high KL divergence from a reference model.
By excluding these high-uncertainty, distribution-shifting tokens from imitation,
{\ourtitle} injects task-specific knowledge while preserving the integrity of the model's pre-trained distribution.
Empirical evaluations on mathematical reasoning benchmarks demonstrate that
{\ourtitle} consistently outperforms standard SFT. 
Further RL fine-tuning from the {\ourtitle} model yields consistently better post-RL performance, indicating improved exploration for the RL stage.
Our codes and datasets are available at \href{https://github.com/MINE-USTC/EKSFT}{https://github.com/MINE-USTC/EKSFT}.
\end{abstract}

\section{Introduction}

%


Recently, the two-stage training pipeline of Supervised Fine-Tuning (SFT) then Reinforcement Learning (RL) 
has demonstrated remarkable capabilities across various downstream applications
of large language models (LLMs)~\citep{dao2025alphamaze, zhi2025medgr, pang2025fevo, yang2025qwen3, openai2023gpt4}, though its limitations are increasingly apparent.
A reasonable viewpoint is that SFT provides a cold-start, as pure RL often yields insufficient positive samples for efficient learning.
However, recent studies suggest that SFT may impair the model's exploration capacity, 
potentially affecting the performance of subsequent RL training~\citep{chen2025sftorrl,zhang2025nemotron,zhang2025onpolicyrl}.
\begin{figure}[t]
    \centering
    \includegraphics[width=1.0\linewidth]{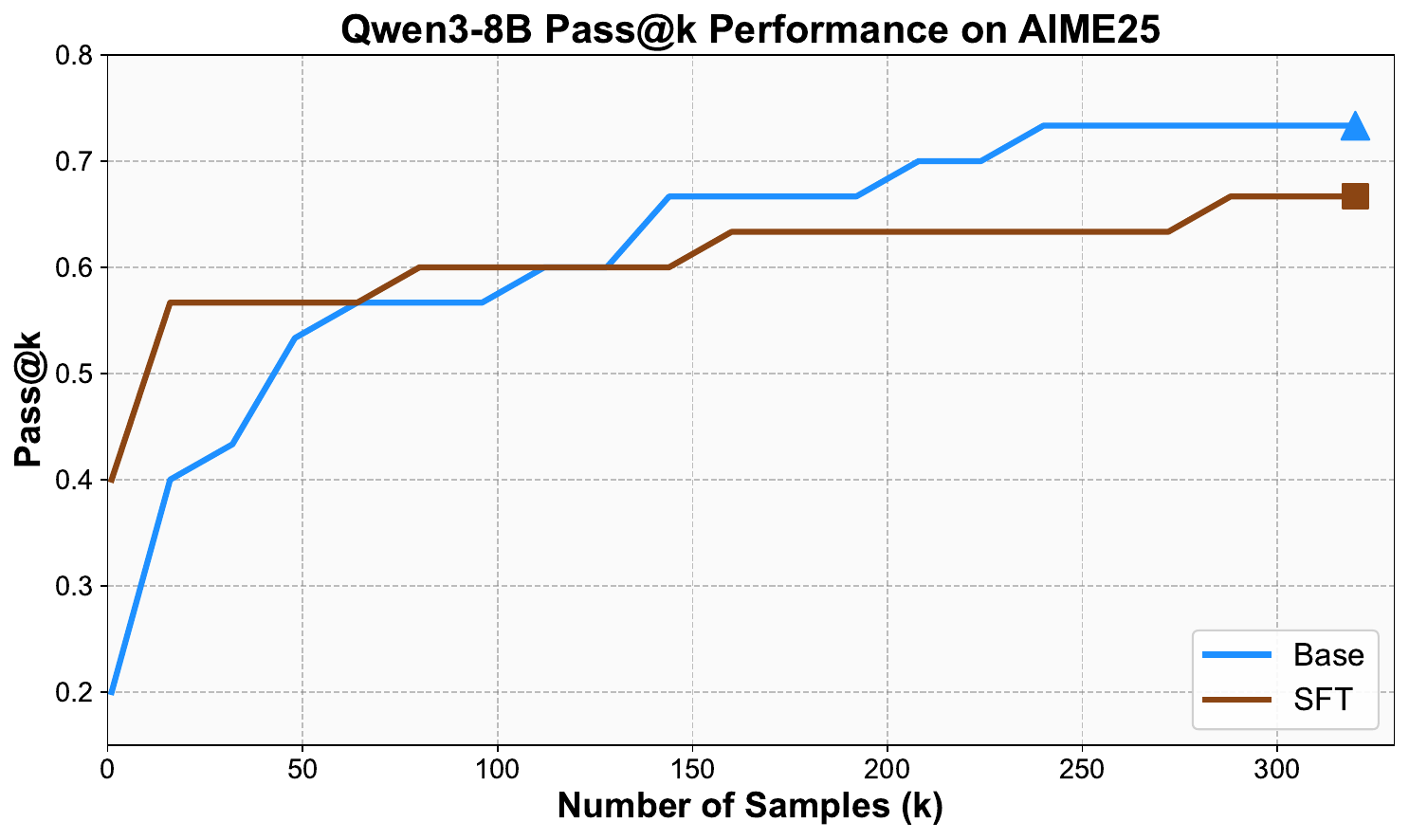}
    \caption{We train Qwen3-8B on the OpenR1 dataset and evaluate the Pass@K performance on AIME25. 
    These results show that SFT models can yield suboptimal performance compared to Base models when K exceeds a threshold, e.g., K=140 as shown.}
    \vspace{-16pt}
    \label{fig:intro}
\end{figure}
This limitation stems from the fact that SFT essentially performs behavior cloning~\citep{chu2025sftmemrlgen, qin2025iwsft, zhu2025psft}
 on expert data, and when the dataset is limited or distributionally misaligned with the pretraining corpus, 
the model may suffer substantial parameter drift, leading to degraded generalization.
Meanwhile, token-by-token imitation tends to sharpen the policy distribution, reducing rollout diversity, 
which is harmful for collecting sufficiently diverse rollouts during the RL phase.
As shown in Figure~\ref{fig:intro}, SFT models exhibit lower Pass@K~\citep{chen2025passk} at large K compared to Base models, 
suggesting that SFT narrows the effective output distribution and reduces sampling diversity.

To address the above issues, existing methods can be categorized into two main research lines. 
A category of methods combines the SFT training objective and RL training objective to leverage the strengths of both paradigms~\citep{zhang2025onpolicyrl,wu2025dft,yan2025luffy,fu2025srft,chen2025bridge,he2025amft}.
For example, DFT~\citep{wu2025dft} reformulates SFT within an RL objective, 
while methods like CHORD~\citep{zhang2025onpolicyrl}, BRIDGE~\citep{chen2025bridge} and AMFT~\citep{he2025amft} use static or dynamic
weighting to balance the two objectives during training.
However, in practice, mixing these two objectives often requires delicate tuning and shows high sensitivity to weighting schedules.
Another line of work focuses on revising the SFT objective to improve generalization in the following RL stage.
For example, PSFT~\citep{zhu2025psft} adopts a clipped surrogate objective to prevent excessive
policy update, while
IW-SFT~\citep{qin2025iwsft} and ASFT~\citep{zhu2025asft} leverage importance weighting~\citep{rubinstein2016simulation} 
to control the distributional shift that maintains the training stability for the RL phase.
However, these methods still supervise all tokens uniformly and rely on global regularization to control deviation.
Moreover, they do not explicitly address entropy collapse~\citep{cui2025entropy}, 
which may diminish the model's exploration capacity~\citep{xie2025exploration}.

To address these challenges, we propose that SFT should regard \emph{activating} task-relevant capabilities
rather than \emph{memorizing} specific content~\citep{chu2025sftmemrlgen, xie2025exploration} 
when SFT serves as an initialization stage for subsequent RL, 
improving the generalization of SFT models and enhancing the exploration for the subsequent RL phase.
Along this line, we propose Entropy-KL Selective Fine-Tuning (\ourtitle),
which selectively masks tokens that exhibit either high entropy or high KL divergence from a reference model,
while high entropy tokens usually indicate greater model uncertainty and high KL-divergence tokens represent deviation from the reference model.
Additionally, 
{\ourtitle} incorporates entropy loss~\citep{zhang2025entropyregularization,cheng2025entropyperspective} and KL divergence 
loss~\citep{li2025gem} constraints to further mitigate entropy collapse and maintain proximity to the pretrained distribution.
By combining selective masking with explicit regularization, 
{\ourtitle} injects task-relevant knowledge while maintaining small parameter drift and mitigating entropy collapse.
The contributions of this paper can be summarized as follows:
\begin{itemize}[itemsep=0pt, topsep=0pt]
    \item We propose a new perspective on the SFT-then-RL paradigm: SFT should regard \emph{activating} task-relevant capabilities
    rather than \emph{memorizing} specific content when SFT serves as an initialization stage for subsequent RL.
    \item We propose {\ourtitle}, a novel approach that refines SFT by selectively masking high-entropy and high-KL-divergence tokens 
    and incorporating entropy and KL regularization loss, thereby enhancing generalization and exploration for the subsequent RL phase.
    \item Empirical evaluations against multiple state-of-the-art (SOTA) baselines across several benchmarks 
    demonstrate the effectiveness and superior performance of our proposed {\ourtitle}.
\end{itemize}

\section{Related Work}

\subsection{Supervised Fine-Tuning}
SFT adapts pre-trained models to downstream tasks by training on task-specific~\citep{chu2025sftmemrlgen, xu2025harnessing, xu2026from} datasets.
It is often used as an initialization stage in the multi-stage post-training pipeline.
However, recent studies have shown that the standard Cross-Entropy (CE) loss is not the best fit for SFT~\citep{li2025gem,xiao2024rethinking},
pointing out that fine-tuning with standard CE loss often significantly reduces diversity.
Moreover, SFT essentially performs behavior cloning~\citep{chu2025sftmemrlgen}, 
and when used as a cold-start for subsequent RL, it may cause distribution sharpening, reducing exploration capacity.
GEM~\citep{li2025gem} proposes a game-theoretic SFT algorithm with reverse Kullback-Leibler (KL) divergence minimization 
to preserve diversity and mitigate overfitting.
IW-SFT~\citep{qin2025iwsft} and ASFT~\citep{zhu2025asft} involve importance-weighted mechanisms to be further generalized.
PSFT~\citep{zhu2025psft} is inspired by TRPO~\citep{schulman2015trpo} to adopt a clipped surrogate objective, 
leaving room for further optimization.
Nonetheless, these approaches still supervise all tokens uniformly with global regularization, 
leading to suboptimal generalization for the subsequent RL phase.

\subsection{Reinforcement Learning for LLM Alignment}
Reinforcement learning (RL) has been widely adopted to 
enhance the alignment of large language models (LLMs) with human preferences~\citep{bai2022traininghh, ouyang2022rlhf}.
Recent advancements show remarkable success in complex reasoning tasks like mathematics and code generation~\citep{deepseekai2024} 
through Reinforcement Learning from Verifiable Reward (RLVR)~\citep{deepseekr12025, RLVR2025survey}.
RLVR leverages programmatic verifiers, such as unit tests~\citep{poznanski2025olmocr} or answer checkers~\citep{yu2025dapo}, 
to provide reward signals to achieve superior performance.
However, RL-based exploration is often constrained by the base model's insufficient positive samples, leading to high computational
expense and learning inefficiency~\citep{wu2025dft,zhu2025psft}.
This limitation motivates an initialization phase that activates the base model's relevant capabilities before the RL phase.

\subsection{Combining Supervised Fine-Tuning and Reinforcement Learning}
To exploit the complementary strengths of SFT and RL,
recent studies have explored their combination~\citep{ouyang2022rlhf,sheng2025hybridflow,liu2025uft}.
SFT-then-RL paradigm~\citep{lambert2024tulu3,liu2025acereasonnemotron} is widely adopted, 
which is a two-stage training pipeline where SFT is first employed to initialize the model,
followed by RL for further optimization, and has been widely adopted in LLM applications, 
e.g., medical and financial domains~\citep{zhi2025medgr,pang2025fevo}.
The concurrent DFT~\citep{wu2025dft} views that SFT's gradient update is equivalent to a policy gradient to rescale the SFT objective.
CHORD~\citep{zhang2025onpolicyrl}, AMFT~\citep{he2025amft}, and BRIDGE~\citep{chen2025bridge} 
unify SFT and RL by designing a joint loss function between SFT and RL objectives with the weighted mechanism.
SRFT~\citep{fu2025srft} integrates both fine-tuning paradigms through entropy-aware weighting mechanisms.
However, joint optimization is often highly sensitive to weighting schedules and can be unstable in practice,
while CE-based imitation may still cause distribution sharpening and reduce exploration.
In contrast, we improve the standard SFT via the selective token masking mechanism and fine-grained regularization,
aiming to enhance the generalization and exploration for the subsequent RL phase.

\section{Preliminary}

\paragraph{Supervised Fine-Tuning.} SFT is a common approach to adapt LLMs to downstream tasks.
Given a supervised dataset $\mathcal{D}=\left\{(x_i,{y_i}^*)\right\}_{i=1}^N$, 
where $x_i$ denotes the prompt and ${y_i}^*=(y_{i,1}, y_{i,2},... )$ denotes the corresponding response of $|{y_i}^*|$ tokens.
The objective is to train the policy model $\pi_\theta$ to 
maximize the likelihood of the response ${y_i}^*$ given the corresponding prompt $x_i$:
\begin{equation}
    \mathcal{L}_{\text{SFT}}(\theta) = - \mathbb{E}_{(x_i,{y_i}^*)\sim \mathcal{D}} \left[ \sum_{t=1}^{|{y_i}^*|} \log \pi_\theta (x_i, y_{i,t}|y_{i,<t}^*) \right].
    \label{eq.1}
\end{equation}
where, $y_{i,<t}^*=(y_{i,1}, y_{i,2},...,y_{i,t-1})$ denotes the previous tokens before the index $t$.

\paragraph{Token-level Entropy.} Token-level entropy quantifies 
the uncertainty of the next-token probability distribution produced by a language model.
Given a prompt $x$ and a language model distribution $\pi_\theta$, the token-level entropy at time step $t$ is defined as:
\begin{equation}
    \begin{split}
        H&(\pi_\theta(\cdot|x,y_{<t})) = \\
        & - \sum_{v\in \mathcal{V}} \pi_\theta(v|x,y_{<t}) \log \pi_\theta(v|x,y_{<t})
    \end{split}
    \label{eq.2}
\end{equation}
where $\mathcal{V}$ denotes the vocabulary and $y_{<t}$ denotes the previous tokens before the index $t$.

\paragraph{Token-level Kullback-Leibler divergence in LLMs.} The KL is 
a statistical divergence that quantifies how one probability distribution differs from another.
Given two language models $\pi_{\theta}$ and $\pi_\mathrm{ref}$, 
the token-level KL divergence from $\pi_{\theta}$ to $\pi_\mathrm{ref}$ is defined as:
\begin{equation}
    \begin{split}
        \text{KL}&(\pi_{\theta}(\cdot|x,y_{<t})||\pi_\mathrm{ref}(\cdot|x,y_{<t})) = \\
        & \sum_{v\in \mathcal{V}} \pi_{\theta}(v|x,y_{<t})\,\log\frac{\pi_{\theta}(v|x,y_{<t})}{\pi_\mathrm{ref}(v|x,y_{<t})}
    \end{split}
    \label{eq.3}
\end{equation}
where $\mathcal{V}$ denotes the vocabulary and $y_{<t}$ denotes the previous tokens before the index $t$.

\section{Methodology}

%
%
\makeatletter
\@ifundefined{sethlcolor}{}{%
  \sethlcolor{white}%
  \soulregister{\ref}{1}%
  \soulregister{\citep}{1}%
  \soulregister{\cite}{1}%
  \soulregister{\emph}{1}%
  \soulregister{\textbf}{1}%
  \soulregister{\ourtitle}{0}%
}
\@ifundefined{mdframed}{%
  \newenvironment{rebuttalblock}{\par}{\par}%
}{%
  \newenvironment{rebuttalblock}{%
    \begin{mdframed}[backgroundcolor=yellow!20, linewidth=0pt,
                      innertopmargin=4pt, innerbottommargin=4pt,
                      innerleftmargin=6pt, innerrightmargin=6pt,
                      skipabove=4pt, skipbelow=4pt]%
  }{%
    \end{mdframed}%
  }%
}
\@ifundefined{hl}{\newcommand{\hl}[1]{{\setlength{\fboxsep}{1pt}\colorbox{yellow}{#1}}}}{}
\makeatother

In this section, we propose {\ourtitle},
aiming to better leverage task-relevant knowledge of LLMs on the expert dataset 
and mitigate the generalization decline and entropy collapse 
by the standard Supervised Fine-Tuning,
which is beneficial for the subsequent RL phase.

\subsection{Token Selection}

Large language models (LLMs) follow an auto-regressive fashion, where the smallest unit during training or inference is the token.
To enhance the model's generalization after training,
recent studies have shown that the effectiveness of designing entropy-preserving mechanisms and incorporating KL-divergence constraints.

\textbf{High-entropy tokens} represent positions where the model is uncertain.
Forcing the model to imitate these uncertain predictions may sharpen the policy distribution and reducing output diversity instead of maintaining the distribution and  activating the task-relevant knowledge.
\textbf{High-KL tokens} indicate positions where the current policy has already diverged significantly from the reference model.
Continuously supervising these tokens would further amplify distributional drift.
Therefore, during the supervised learning, we propose to mask tokens that exhibit high entropy or high KL divergence, allowing the model to activate the task-relevant knowledge while preserving its exploration capacity and proximity to the pretrained distribution.
\hl{A formal token-level gradient analysis is provided in Appendix~\mbox{\ref{app:theory}}.}

To avoid setting specific thresholds for determining the category of a token, we introduce a top‑k ratio strategy, which ranks tokens within a given set.
Given a policy model $\pi_\theta$, a reference model $\pi_\mathrm{ref}$, the Top-K ratio $\rho$ ($\rho \in (0,1]$) and sequential tokens $T=\left\{t_1, t_2, ..., t_{|T|}\right\}$,
we can easily calculate the token-level entropy at the $m$-th token $t_m$:
\begin{equation}
    {H}(t_m) =
    - \sum_{v\in \mathcal{V}} \pi_\theta(v|x,y_{<t}) \log \pi_\theta(v|x,y_{<t}),
    \label{eq.4}
\end{equation}
where $\mathcal{V}$ denotes the vocabulary of the policy model.
Then, we obtain all tokens entropy in $T$:
\begin{equation}
    {H}(T) = \left\{{H}(t_1), {H}(t_2), ..., {H}(t_{|T|})\right\}.
    \label{eq.5}
\end{equation}
And the rank of the $m$-th token by entropy can be formulated as:
\begin{equation} \label{eq.6}
    r(t_m, H(T)) =
    \sum_{i=1}^{|T|}
    \mathbf{I}\left[H(t_i) \ge H(t_m)\right],
\end{equation}
where $\mathbf{I}$ is the indicator function.
Subsequently, we select the top‑k ratio tokens, defined as follows:
\begin{equation}\label{eq.7}
\operatorname{Top\text{-}K}(H(T),\rho)
= \left\{t_i| r(t_i, H(T)) \le k \right\}, 
\end{equation}
where $k = \left\lceil \rho \cdot |T| \right\rceil$ denotes the number of selected elements.

Beyond entropy, we found that tokens with large KL divergence have a pronounced effect on the drift of model parameters. 
Similar to token-level entropy computation, the KL divergence at the $m$-th token $t_m$ can be calculated as:
\begin{equation}\label{eq.8}
    \operatorname{KL}(t_m) = 
    \sum_{v\in \mathcal{V}} \pi_{\theta}(v|x,y_{<t})\,\log\frac{\pi_{\theta}(v|x,y_{<t})}{\pi_{\mathrm{ref}}(v|x,y_{<t})}.
\end{equation}
Then, all tokens KL divergence in $T$ are as follows:
\begin{equation}
    \operatorname{KL}(T) = \left\{\operatorname{KL}(t_1), \operatorname{KL}(t_2), ..., \operatorname{KL}(t_{|T|})\right\}.
    \label{eq.9}
\end{equation}
Following the same steps in Equation~\ref{eq.6} and Equation~\ref{eq.7},
we can identify the tokens with the highest Top-K ratio among the KL divergence tokens,
given by:
\begin{equation}\label{eq.10}
\operatorname{Top\text{-}K}(\operatorname{KL}(T),\rho)
= \left\{t_i \,\middle|\, r\!\left(t_i,\operatorname{KL}(T)\right)\le k \right\}. 
\end{equation}

\paragraph{Combining Entropy and KL Selection.}
Given the two Top-K sets, we adopt a \emph{union} strategy to construct the final selected token set:
\begin{equation}\label{eq.11}
\begin{split}
\mathcal{M}_{\mathrm{KL}} &= \operatorname{Top\text{-}K}(\operatorname{KL}(T),\rho);
\\
\mathcal{M}_{H} &= \operatorname{Top\text{-}K}(H(T),\rho);
\\
\mathcal{M} &= \mathcal{M}_{\text{KL}} \cup \mathcal{M}_{H}.
\end{split}
\end{equation}
This union covers tokens that are either high entropy or high KL, ensuring that our selective masking mitigates both entropy collapse and distributional shift during supervised learning.
\hl{The two sets are largely complementary (average IoU \mbox{$\approx 0.50$}); see Appendix~\mbox{\ref{app:iou}} for details.}

\begin{algorithm*}[!htbp]
\caption{EKSFT: Entropy-KL Selective Fine-Tuning}
\label{alg:eksft}
\small
\KwIn{
Policy model $\pi_\theta$, Reference model $\pi_{\mathrm{ref}}$, Dataset $\mathcal{D}$, Top-K ratio $\rho \in (0,1]$, Coefficients $\lambda_H, \lambda_{\mathrm{KL}}$, \\
Learning rate $\eta$
}

\For{each epoch}{
    Sample a batch $\mathcal{B}=\{(x_{i},y_{i})\}_{i=1}^{B} \sim \mathcal{D}$\;

    Collect all valid tokens 
    $T \gets \{t_1,t_2,\dots,t_{|T|}\}$\;

    \For{each token $t \in T$}{
        $
        H(t) \gets -\sum_{v\in\mathcal{V}}
        \pi_\theta(v|x,y_{<t}) \log \pi_\theta(v|x,y_{<t})
        $\;
        $
        \mathrm{KL}(t) \gets
        \sum_{v\in\mathcal{V}}
        \pi_\theta(v|x,y_{<t})
        \log \frac{\pi_\theta(v|x,y_{<t})}{\pi_{\mathrm{ref}}(v|x,y_{<t})}
        $\;
    }

    Determine masked token number: $k \gets \left\lceil \rho \cdot |T| \right\rceil$\;
    $\mathcal{M}_{H} \gets \operatorname{Top\text{-}K}(H(T), \rho)$;
    $\mathcal{M}_{\mathrm{KL}} \gets \operatorname{Top\text{-}K}(\mathrm{KL}(T), \rho)$\;

    Union masked token set: 
    $\mathcal{M} \gets \mathcal{M}_{H} \cup \mathcal{M}_{\mathrm{KL}}$\;

    Compute masked cross-entropy loss: 
    $
    \mathcal{L}_{\mathrm{CE}}^{\mathrm{mask}}
    \gets
    -\frac{1}{|\,\, \overline{\!\!\mathcal{M}\!} \,|}
    \sum_{t \notin \mathcal{M}}
    \log \pi_\theta(y_t|x_{<t})
    $\;

    Compute masked entropy regularization:
    $
    \mathcal{L}_{H}^{\mathrm{mask}}
    \gets
    \frac{1}{|\mathcal{M}|}
    \sum_{t \in \mathcal{M}} H(\pi_\theta(\cdot|x,y_{<t}))
    $\;

    Compute masked KL regularization:
    $
    \mathcal{L}_{\mathrm{KL}}^{\mathrm{mask}}
    \gets
    \frac{1}{|\mathcal{M}|}
    \sum_{t \in \mathcal{M}}
    \mathrm{KL}(\pi_\theta(\cdot|x,y_{<t}) \| \pi_{\mathrm{ref}}(\cdot|x,y_{<t}))
    $\;

    Combine objectives: 
    $
    \mathcal{L}_{\mathrm{EKSFT}}
    \gets
    \mathcal{L}_{\mathrm{CE}}^{\mathrm{mask}}
    - \lambda_H \mathcal{L}_{H}^{\mathrm{mask}}
    + \lambda_{\mathrm{KL}} \mathcal{L}_{\mathrm{KL}}^{\mathrm{mask}}
    $\;

    $\theta \gets \theta - \eta \nabla_\theta \mathcal{L}_{\mathrm{EKSFT}}$\;
}
\KwOut{Fine-tuned policy model $\pi_\theta$}

\end{algorithm*}

\subsection{Regularization Objectives}

\paragraph{KL-divergence Regularization.}
To minimize the fine-tuned model and the base policy model, 
it is common for a training paradigm to add the KL-divergence regularization in the total loss function.
Although we have masked the high entropy and KL tokens during the training phase, we found that these tokens still suffer from entropy decline and KL increase by the gradient backpropagation in practice.
Therefore, following prior work, we employ a KL-divergence regularization on the masked tokens to prevent undesirable drift, without altering the core design of \ourtitle.
The masked KL-divergence regularization is defined as:
\begin{equation}\label{eq.12}
\begin{aligned}
\mathcal{L}_{\mathrm{KL}}^{\mathrm{mask}}&(\theta)
= \frac{1}{|\mathcal{M}|}
\\
&
\sum_{t \in \mathcal{M}}
\mathrm{KL}\!\left(
\pi_{\theta}(\cdot \mid x, y_{<t})
\,\big\|\,
\pi_{\mathrm{ref}}(\cdot \mid x, y_{<t})
\right)
\end{aligned}
\end{equation}
where $t$, $\pi_\theta$ and $\pi_{\text{ref}}$ represent the masked token in $\mathcal{M}$, the policy model and the base model.

\paragraph{Entropy Regularization.} 
Aligning with the intuition behind KL regularization, to mitigate the entropy collapse on the fine-tuned model,
{\ourtitle} also introduces an entropy regularization on the masked tokens to control the entropy decline. 
This entropy regularization can be formulated as:
\begin{equation}\label{eq.13}
    \mathcal{L}_{H}^{\mathrm{mask}}(\theta) = \frac{1}{|\mathcal{M}|} \sum_{t \in \mathcal{M}}{H}(\pi_\theta(\cdot|x,y_{<t})).
\end{equation}

\subsection{Entropy-KL Selective Fine-Tuning}

The key insight of Entropy-KL Selective Fine-Tuning (\ourtitle) is to activate the task-relevant knowledge while preserving the generalization of the fine-tuned model.
To achieve this goal, {\ourtitle} modifies the standard Cross-Entropy loss by masking Top-K of the high entropy and high KL-divergence tokens.
The main loss function can be reformulated as:
\begin{equation}\label{eq.14}
    \mathcal{L}_{\mathrm{CE}}^{\mathrm{mask}}(\theta) =- \frac{1}{|\,\, \overline{\!\!\mathcal{M}\!} \,|}
\sum_{t \notin \mathcal{M}} \log  \pi_\theta(y_t \mid x_{<t}),
\end{equation}
where $\,\, \overline{\!\!\mathcal{M}\!} \,$ denotes the complement of $\mathcal{M}$.

In addition, to further control these masked tokens' entropy and KL divergence, we supply two weighted regularization: \textbf{Entropy Regularization} and \textbf{KL Regularization}.
Specifically, we define the {\ourtitle} loss by combining the Equation~\ref{eq.12}, Equation~\ref{eq.13} and Equation~\ref{eq.14} as:
\begin{equation}\label{eq.15}
    \mathcal{L}_{\mathrm{\ourtitle}}= \mathcal{L}_{\mathrm{CE}}^{\mathrm{mask}} - \lambda_{H}\mathcal{L}_{H}^{\mathrm{mask}}
    +\lambda_{\mathrm{KL}}\mathcal{L}_{\mathrm{KL}}^{\mathrm{mask}},
\end{equation}
where $\lambda_{H}$ and $\lambda_{\mathrm{KL}}$ represent the hyperparameter weight of entropy regularization and KL regularization, respectively.
\hl{Notably, on masked tokens this yields a label-free gradient that drops the one-hot term \mbox{$-\nabla\log\pi_\theta(y_t)$} of standard SFT, mitigating entropy collapse and drift without negating capability activation. The derivation is given in Appendix~\mbox{\ref{app:theory}}.}

\subsection{Implementation details}
As shown in Algorithm~\ref{alg:eksft}, {\ourtitle} first calculates the token-level entropy and KL divergence in a batch $\mathcal{B}$ by using the Equation~\ref{eq.2} and Equation~\ref{eq.3}.
And then we determine the current quantity $k$ of masked tokens by multiplying the Top-K ratio $\rho$ and the number of the valid tokens, which do not include system prompt tokens.
Subsequently, we can obtain two index sets $\mathcal{M}_{H}$ and $\mathcal{M}_{\text{KL}}$ by taking the Top-K operation for all token-level entropy and KL divergence, respectively.
The final mask set $\mathcal{M}$ aggregates tokens that satisfy either high entropy or high KL divergence.
Based on this mask set, we can compute a masked cross-entropy loss on the complement positions $\,\, \overline{\!\!\mathcal{M}\!} \,$ by applying the Equation~\ref{eq.14}, together with entropy and KL regularization evaluated on the masked positions by individually using Equation~\ref{eq.13} and Equation~\ref{eq.12}.
By combining the masked loss with two weighted regularization, we
can obtain the final training objective for \ourtitle.

\section{Experiments}

\begin{table*}[t]
\centering
\small
\renewcommand{\arraystretch}{1.2}

\resizebox{\textwidth}{!}{%
\begin{tabular}{l|cc cc cc cc | cc}
\hline
\multirow{2}{*}{\textbf{Method}} & \multicolumn{2}{c}{\textbf{AIME}} & \multicolumn{2}{c}{\textbf{AIME25}} & \multicolumn{2}{c}{\textbf{AMC}} & \multicolumn{2}{c|}{\textbf{HMMT25}} & \multicolumn{2}{c}{\textbf{Avg}} \\
 & pass@1 & pass@32 & pass@1 & pass@32 & pass@1 & pass@32 & pass@1 & pass@32 & pass@1 & pass@32 \\
\hline
\multicolumn{11}{c}{\textbf{\textit{Qwen3-4B}}} \\ 
\hline
Base & 25.6 & 60.0 & 21.3 & 46.7 & 60.5 & 86.7 & 11.5 & 30.0 & 29.7 & 55.8 \\
SFT & \textbf{46.3} & \textbf{73.3} & \underline{32.1} & \underline{56.7} & 68.7 & 86.7 & \underline{16.9} & \underline{36.7} & \underline{41.0} & 63.3 \\
DFT & 36.5 & 70.0 & 29.5 & \underline{56.7} & 63.1 & 85.5 & 12.7 & 26.7 & 35.5 & 59.7 \\
IW-SFT & 45.3 & \textbf{73.3} & 31.9 & \underline{56.7} & \textbf{69.2} & \underline{89.1} & 16.8 & \textbf{50.0} & 40.8 & \underline{67.2} \\
PSFT & 40.9 & 66.7 & 29.8 & 46.7 & 68.2 & 83.1 & 15.5 & 33.3 & 38.6 & 57.4 \\ 
{\ourtitle} (Ours) & \underline{45.7} & \textbf{73.3} & \textbf{33.4} & \textbf{60.0} & \underline{68.9} & \textbf{90.4} & \textbf{18.5} & \textbf{50.0} & \textbf{41.7} & \textbf{68.4} \\
\hline
\multicolumn{11}{c}{\textbf{\textit{Qwen3-8B}}} \\ 
\hline
Base & 27.1 & 60.0 & 22.4 & 53.3 & 62.0 & 89.2 & 13.1 & \underline{40.0} & 31.2 & 60.6 \\
SFT & 41.4 & \textbf{83.3} & 27.8 & 56.7 & \underline{70.4} & \textbf{91.6} & 16.9 & \textbf{50.0} & 39.1 & \underline{70.4} \\
DFT & 37.3 & 73.3 & 29.7 & 53.3 & 63.7 & 83.1 & 14.2 & \underline{40.0} & 36.2 & 62.4 \\
IW-SFT & \underline{43.9} & 76.7 & \underline{32.3} & \underline{60.0} & \underline{70.4} & 86.7 & \underline{17.4} & \textbf{50.0} & \underline{41.0} & 68.4 \\
PSFT & 29.0 & 66.7 & 20.5 & 53.3 & 62.1 & 88.0 & 12.9 & 36.7 & 31.1 & 61.2 \\ 
{\ourtitle} (Ours) & \textbf{47.2} & \underline{80.0} & \textbf{34.8} & \textbf{66.7} & \textbf{70.8} & \underline{90.4} & \textbf{19.6} & \textbf{50.0} & \textbf{43.1} & \textbf{71.8} \\
\hline
\end{tabular}%
}
\caption{\textbf{Overall performance comparison (Stage 1: without RL).} We compared the pass@1 and pass@32 performance between different methods using the Qwen3-4B and Qwen3-8B models on four mathematical reasoning benchmarks.}
\vspace{-0.5cm}
\label{tab:stage1}
\end{table*}

\subsection{Experiment Setup}
\paragraph{Training Datasets.}
We performed all training methods based on OpenR1-Math-46k-8192~\citep{yan2025luffy}, which contains 46k prompts and off-policy reasoning traces by filtering out the wrong and longer than 8192 tokens' generations in OpenR1-Math-220k~\citep{openr1}.
In our setting, we sample 3k samples for the SFT and its variants and the remaining 43k samples for the subsequent RL phase, ensuring no overlap.
\vspace{-0.1cm}
\paragraph{Models.} We conducted fine-tuning experiments using two models from Qwen3 family~\citep{yang2025qwen3},
Qwen3-4B and Qwen3-8B.
Both models have demonstrated robust capabilities~\citep{Qwen3Guard, qwen3omini, qwen3emb, qwen3think} across various tasks, showing strong performance whether fine-tuned or used directly.
\vspace{-0.1cm}
\paragraph{Evaluation benchmarks.}
We conducted a comprehensive evaluation on several widely-adopted mathematical reasoning benchmarks, including AIME24~\citep{li2024numinamath}, AIME25~\citep{li2024numinamath}, AMC~\citep{li2024numinamath}
and HMMT25~\citep{balunovic_srimatharena_2025}.
We reported pass@1 to evaluate performance and pass@32 to evaluate
the boundaries of LLM capabilities~\citep{yue2025passk,chen2025passk,brown2024passk}.
\vspace{-0.1cm}
\paragraph{Baselines Methods.}
We compared the proposed {\ourtitle} with a comprehensive set of baselines, including three categories: base model (without training), SFT-variants, and combined SFT with RL methods.
For the base model, we consider the Original Model, which is the original Qwen3-4B/Qwen3-8B model.
For SFT-variants methods, we consider
(1) Standard SFT trains the base model using the standard CE loss function on the SFT dataset.
(2) PSFT~\citep{zhu2025psft} is a trust-region–inspired supervised fine-tuning objective that clips the policy ratio to constrain policy drift.
For combined SFT with RL methods, we consider
(1) IW-SFT~\citep{qin2025iwsft} is an importance-weighted variant that interprets SFT on curated data as a lower bound on a sparse-reward RL objective.
(2) DFT~\citep{wu2025dft} rescales the SFT objective by using a simple
probability-based reweighting mechanism.

\vspace{-0.1cm}
\paragraph{Training Details.} 
For a fair comparison, all training methods use the same experimental data and hyperparameter settings.
To further compare the cold-start performance of different methods,
we initialize the second-stage reinforcement learning with models trained by each method under uniform configurations.
All hyperparameters remain consistent during RL training.
Details are in Appendix~\ref{app:imple_details}.

\begin{table*}[t]
\centering
\small
\renewcommand{\arraystretch}{1.2}

\resizebox{\textwidth}{!}{%
\begin{tabular}{l|cc cc cc cc | cc}
\hline
\multirow{2}{*}{\textbf{Method(+DAPO)}} & \multicolumn{2}{c}{\textbf{AIME}} & \multicolumn{2}{c}{\textbf{AIME25}} & \multicolumn{2}{c}{\textbf{AMC}} & \multicolumn{2}{c|}{\textbf{HMMT25}} & \multicolumn{2}{c}{\textbf{Avg}} \\

 & pass@1 & pass@32 & pass@1 & pass@32 & pass@1 & pass@32 & pass@1 & pass@32 & pass@1 & pass@32 \\
\hline
\multicolumn{11}{c}{\textbf{\textit{Qwen3-4B}}} \\ 
\hline
Base & 43.4 & 73.3 & 37.6 & \underline{66.7} & 74.9 & 91.6 & 17.8 & 33.3 & 43.4 & 66.2 \\
SFT & \underline{48.1} & \underline{80.0} & \underline{39.5} & \underline{66.7} & \underline{76.4} & \underline{94.0} & \underline{19.4} & \underline{40.0} & 45.8 & \underline{70.2} \\
DFT & 39.3 & 73.3 & 31.5 & 63.3 & 72.1 & 90.4 & 13.7 & 33.3 & 39.1 & 65.1 \\
IW-SFT & 46.0 & 66.7 & 36.3 & 63.3 & 76.2 & 91.6 & \textbf{21.4} & \textbf{50.0} & 45.0 & 67.9 \\
PSFT & 46.3 & 76.7 & 39.3 & 60.0 & \textbf{78.2} & 90.3 & 20.0 & 46.7 & \underline{45.9} & 68.4 \\ 
{\ourtitle} (Ours) & \textbf{48.3} & \textbf{83.3} & \textbf{41.6} & \textbf{73.3} & \textbf{78.2} & \textbf{96.4} & \textbf{21.4} & \textbf{50.0} & \textbf{47.4} & \textbf{75.8} \\
\hline
\multicolumn{11}{c}{\textbf{\textit{Qwen3-8B}}} \\ 
\hline
Base & 50.6 & 73.3 & 37.4 & 63.3 & 78.9 & 90.4 & 19.6 & \textbf{53.3} & 46.6 & 70.1 \\
SFT & 52.6 & 80.0 & 40.0 & 66.7 & 80.5 & 93.9 & \underline{20.0} & \underline{50.0} & 48.3 & 72.7 \\
DFT & 39.5 & 76.7 & 31.3 & 63.3 & 71.7 & 90.4 & 16.9 & 33.3 & 39.9 & 65.9 \\
IW-SFT & \underline{54.6} & \textbf{83.3} & \underline{41.0} & \underline{73.3} & \underline{80.8} & \underline{94.0} & \underline{20.0} & \underline{50.0} & \underline{49.1} & \underline{75.1} \\
PSFT & 52.8 & 73.3 & 38.9 & 63.3 & \underline{80.8} & 91.6 & 18.6 & \underline{50.0} & 47.8 & 69.6 \\ 
{\ourtitle} (Ours) & \textbf{56.5} & \textbf{83.3} & \textbf{42.5} & \textbf{83.3} & \textbf{82.1} & \textbf{96.4} & \textbf{23.3} & \underline{50.0} & \textbf{51.1} & \textbf{78.3} \\
\hline
\end{tabular}%
}
\caption{\textbf{Overall performance comparison (Stage 2: after RL).} We compared the pass@1 and pass@32 between different methods using the Qwen3-4B and Qwen3-8B models on four mathematical reasoning benchmarks.}
\vspace{-0.5cm}
\label{tab:stage2}
\end{table*}

\subsection{Main Results}
Our main results can be divided into two parts according to the training stages. 
In the first stage, we performed supervised learning on the 3k dataset. 
In the second stage, starting from the supervised fine-tuned models, 
we further continued training with reinforcement learning on the remaining 43k dataset. 
The results from the two stages are summarized in Table~\ref{tab:stage1} and Table~\ref{tab:stage2}, respectively.
\vspace{-0.1cm}
\paragraph{Performance after the supervised learning.} 
Across both model scales, \textbf{\ourtitle} generally achieves higher performance (pass@1) and stronger exploration capability (pass@32) than baselines.
For the \textbf{Qwen3-4B model}, compared to the standard SFT, {\ourtitle} yields an average \textbf{+0.7\%} improvement in pass@1 and \textbf{+5.1\%} improvement in pass@32,
which demonstrates that the masking mechanism of high entropy and high KL divergence tokens in {\ourtitle} not only enhances its downstream performance (pass@1), but also improves its exploratory capability (pass@32) on downstream tasks.
In addition, compared to the competitive baseline IW-SFT, {\ourtitle} still yields an average \textbf{+0.9\%} improvement in pass@1 and \textbf{+1.2\%} improvement in pass@32.
A similar trend is observed for the \textbf{Qwen3-8B model}, 
where {\ourtitle} reaches the highest average scores: \textbf{43.1\%} in pass@1 and \textbf{71.8\%} in pass@32.
The best baselines for both pass@1 and pass@32 are IW-SFT and standard SFT.
{\ourtitle} still outperforms IW-SFT by \textbf{+2.1\%} in pass@1 and outperforms standard SFT by \textbf{+1.4\%} in pass@32.

We found that {\ourtitle} gains more improvement on HMMT25, which is a comprehensive mathematical challenging benchmark, 
suggesting that avoiding token-level imitation on the high entropy and high KL divergence tokens is essential for supervised learning on the complex downstream tasks.
Overall, the results indicate that {\ourtitle} can enhance the reasoning performance and exploration diversity, 
showing that {\ourtitle} provides an effective cold start for the subsequent RL phase.
In addition, we also compared the parameters drift by calculating the changes in model parameters before and after training, shown in Appendix~\ref{app:parameter_drift}.
\begin{table*}[t]
\centering
\small
\renewcommand{\arraystretch}{1.2}

\resizebox{\textwidth}{!}{%
\begin{tabular}{l|cc cc | cc cc}
\hline
\multirow{2}{*}{\textbf{Model}} & \multicolumn{2}{c}{\textbf{AIME}} & \multicolumn{2}{c|}{\textbf{AIME25}} & \multicolumn{2}{c}{\textbf{AMC}} & \multicolumn{2}{c}{\textbf{HMMT25}} \\
 & pass@1 & pass@32 & pass@1 & pass@32 & pass@1 & pass@32 & pass@1 & pass@32 \\
\hline
\multicolumn{9}{c}{\textbf{\textit{Qwen3-4B}}} \\ 
\hline
\ourtitle & \textbf{45.7} & \textbf{73.3} & \textbf{33.4} & \textbf{60.0} & \textbf{68.9} & \textbf{90.4} & \textbf{18.5} & \textbf{50.0} \\
\hline
w/o Entropy Regularization & 42.6 & 66.7 & 31.1 & 53.3 & 68.7 & 86.7 & 15.8 & 36.7 \\
w/o KL Regularization & 41.9 & 70.0 & 30.8 & 56.7 & 65.2 & 86.7 & 16.5 & 46.7 \\
w/o Entropy \& KL Regularization & 41.7 & 70.0 & 31.6 & 50.0 & 67.8 & 86.7 & 16.4 & 43.3 \\
\hline
\multicolumn{9}{c}{\textbf{\textit{Qwen3-8B}}} \\ 
\hline
\ourtitle & \textbf{47.2} & \textbf{80.0} & \textbf{34.8} & \textbf{66.7} & \textbf{70.8} & \textbf{90.4} & \textbf{19.6} & \textbf{50.0} \\
\hline
w/o Entropy Regularization & 45.4 & 76.7 & 32.1 & 53.3 & 69.2 & 88.0 & 16.3 & 40.0 \\
w/o KL Regularization & 42.2 & 73.3 & 32.3 & 56.7 & 67.6 & 86.7 & 18.7 & 50.0 \\
w/o Entropy \& KL Regularization & 45.0 & 73.3 & 32.3 & 60.0 & 70.2 & 87.9 & 17.7 & 46.7 \\
\hline
\end{tabular}
}

\caption{Ablation study of different regularization.}
\vspace{-0.5cm}
\label{tab:ablation}
\end{table*}
\vspace{-10pt}
\paragraph{Performance after the RL training.} To verify the effectiveness of different methods as the
cold start point of RL, we further continued training with RL on the remaining 43k dataset.
As shown in Table~\ref{tab:stage2}, {\ourtitle}+DAPO still achieves the best overall performance under both Qwen3-4B and Qwen3-8B, consistently leading on Avg pass@1 and Avg pass@32.
For the \textbf{Qwen3-4B model}, {\ourtitle} reaches \textbf{47.4\%} in pass@1 and \textbf{75.8\%} in pass@32, outperforming PSFT by \textbf{1.5\%} in pass@1 and SFT by \textbf{5.6\%} in pass@32.
For the \textbf{Qwen3-8B model}, {\ourtitle} reaches \textbf{51.1\%} in pass@1 and \textbf{78.3\%} in pass@32, outperforming IW-SFT by \textbf{2.0\%} in pass@1 and by \textbf{3.2\%} in pass@32.
In detail, we observed a positive correlation between pass@1 and pass@32 during the RL training phase: models with higher pass@32 consistently achieve higher pass@1.
For instance, when using Qwen3-4B as the base model, {\ourtitle}+DAPO and SFT+DAPO achieve the best and second-best pass@32 scores, and correspondingly, they also attain the best and second-best pass@1 scores. The same trend holds for the Qwen3-8B model.
 
Therefore, as a cold start point of RL, a key challenge is to \textbf{enhance the model's exploratory ability} while maintaining its accuracy, which is crucial for improving the subsequent RL phase.
By masking tokens with high uncertainty and those that significantly deviate from the model during supervised fine-tuning, our approach {\ourtitle} simultaneously enhances the model's downstream performance and its exploratory capability. Consequently, it achieves superior results in the subsequent reinforcement learning stage compared to other baselines.
\hl{Beyond mathematical reasoning, we validated \mbox{\ourtitle} on the tool-use benchmark BFCL~\mbox{\citep{patil2025bfcl}} after training on AceBench~\mbox{\citep{chen2025acebench}}, where \mbox{\ourtitle} consistently outperforms SFT and DFT on both Qwen3-4B and Qwen3-8B; full results are in Appendix~\mbox{\ref{app:bfcl}}.}
\vspace{-0.1cm}
\paragraph{Details pass@{k} analysis.} To examine the exploration boundary of different supervised learning methods, we increased \textit{k} in the pass@{k} metric.
We chose AIME25 as the evaluation dataset and reported the trend between different approaches on Qwen3-8B by extending \textit{k} up to 320.

\begin{figure}[htbp]
    \centering
    \includegraphics[width=1.0\linewidth]{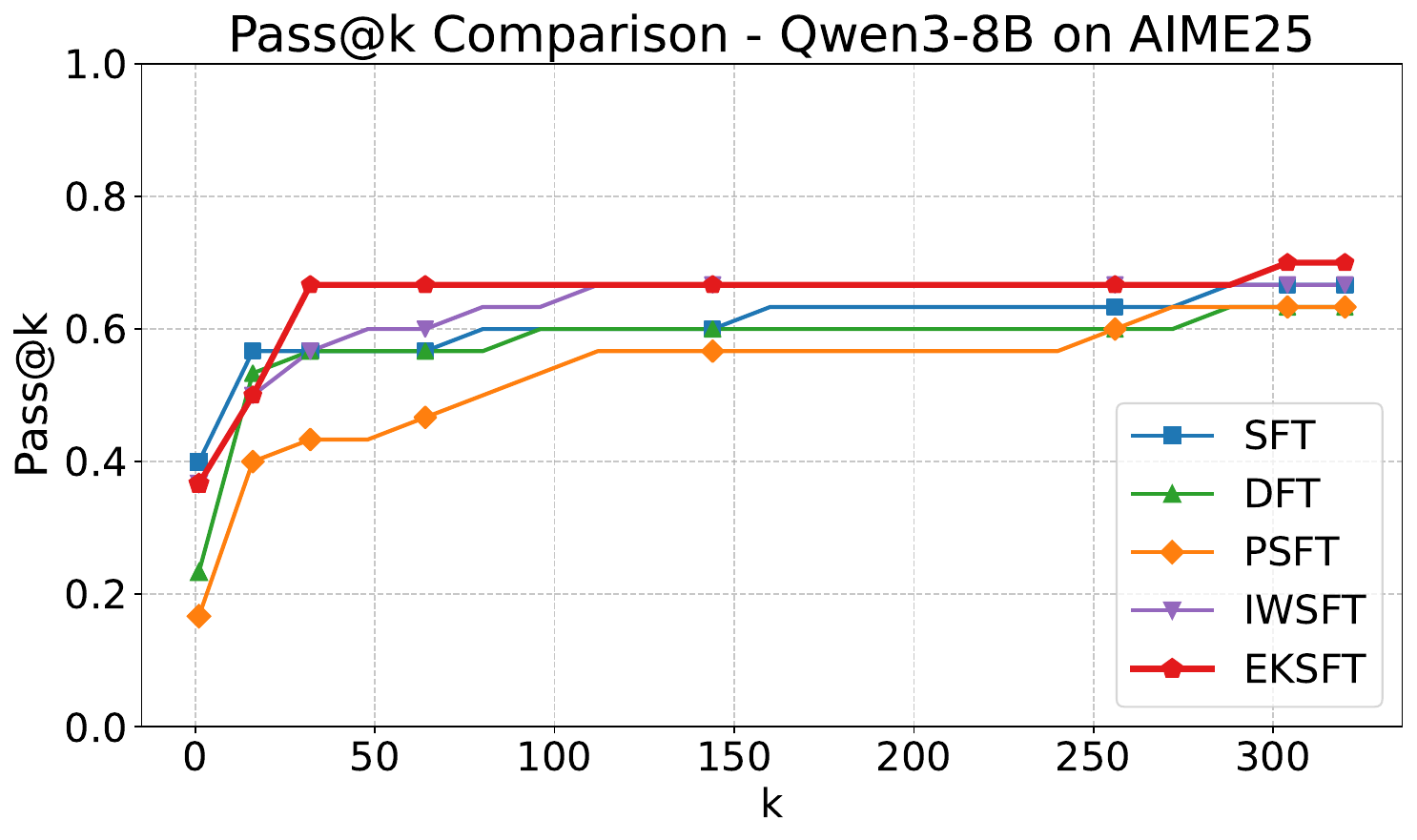}
    \vspace{-0.8cm}
    \caption{We compared the pass@k on AIME25 based on Qwen3-8B model.}
    \label{fig:passk}
\end{figure}

As shown in Figure~\ref{fig:passk}, {\ourtitle} consistently achieved
higher pass@k performance when $k$ is greater than 32,
indicating stronger exploration and reasoning diversity.
In addition, we found that {\ourtitle} exhibits a steeper performance curve, achieving higher pass@k faster, showing that our approach increases the sampling probability of the correct solution under
constrained budgets ($k \le 32$).
In the high-\textit{k} regime, {\ourtitle} could reach a higher performance than all baselines, indicating a richer and more effective distribution of correct solutions.
More results are shown in Appendix~\ref{app:more_results}.

\vspace{-16pt}
\subsection{Ablation Study}
In this section, we performed an ablation study to examine the contributions of {\ourtitle} components, 
including entropy regularization and KL-divergence regularization as summarized in
Table~\ref{tab:ablation}. 
Overall, each regularization is important, but they contribute in different ways.
\textbf{Removing entropy regularization causes the largest degradation on pass@32}, with the most dramatic drop appearing on HMMT25: \textbf{-13.3\%} on Qwen3-4B and \textbf{-10.0\%} on Qwen3-8B, which is consistent that the model may produce more repetitive but wrong path, reducing the chance that multiple samples explore diverse solution paths with the decline of entropy.
In contrast, \textbf{removing KL-divergence degrades the pass@1}, 
especially on AIME: \textbf{-3.8\%} on Qwen3-4B and \textbf{-5.0\%} on Qwen3-8B, indicating that KL-divergence acts like an anchor term, which stabilizes training and preserves the useful reasoning patterns learned during supervised learning, improving pass@1 correctness. 
Interestingly, removing both regularization techniques is not uniformly the worst across all datasets, suggesting
that entropy regularization and KL-divergence regularization are not simply additive, trading off exploration and constraint during training.
In addition,
we further compared {\ourtitle} with a global-regularized SFT baseline and found that {\ourtitle} consistently performs better across all benchmarks, highlighting the importance of the selective masking mechanism. Detailed results are provided in Appendix~\ref{app:masking_mech}. 
\hl{To further verify that the gain stems from \mbox{\emph{selectivity}} rather than removing supervision, we included a random-masking baseline in Appendix~\mbox{\ref{app:random_mask}}, and analyzed the sensitivity of \mbox{\ourtitle} to the masking ratio \mbox{$\rho$} in Appendix~\mbox{\ref{app:ratio}}.}

\vspace{-0.1cm}
\section{Conclusions}

\vspace{-0.2cm}
In this paper, we propose Entropy-KL Selective Fine-Tuning (\ourtitle), a simple yet effective modification to supervised fine-tuning for the widely used SFT-then-RL post-training paradigm,
which masks high entropy or high KL-divergence tokens from token-level imitation in the cross-entropy objective,
mitigating entropy collapse or distributional drift during supervised learning.
{\ourtitle} also applies entropy and KL regularization on these masked tokens to maintain diversity and stay close to the pretrained distribution, while activating the task-relevant knowledge from the remaining tokens.
Extensive experiments on mathematical reasoning benchmarks with Qwen3-4B/8B demonstrate that {\ourtitle} consistently outperforms standard SFT and strong SFT variants, and also yields stronger post-RL performance when used as the initialization for the subsequent RL stage, 
suggesting that the masking mechanism in our approach provides a more robust and exploration-friendly initialization for reinforcement learning.



\clearpage

\section*{Limitations}

\paragraph{Lack of attempts on more datasets.} 
In our experiment setting, we only train the policy model on OpenR1-Math-46k-8192, lacking consideration for the impact of different training datasets, such as the DeepscaleR dataset.

\paragraph{Lack of attempts on more model scales.}
Due to high computational cost, our
experiments are only trained on Qwen3-4B and Qwen-8B models.
Future work should focus on different backbones, such as the LLaMA series or model scales larger than 32B parameters.

\section*{Ethics Statement}

We affirm our commitment to the ethical conduct of this research and provide the following assurances:
\begin{itemize}
    \item This research was conducted in strict adherence to the highest ethical standards, and all findings have been reported with integrity, ensuring clarity and accuracy throughout our communications.
    \item Our study strictly avoids the use of sensitive or confidential data, ensuring that all materials are appropriate for public dissemination.
    \item The datasets employed in our experiments are sourced from publicly accessible and peer-reviewed scientific resources, insuring transparency and reliability.
    \item We provided a detailed description of the dataset characteristics and the hyper-parameter settings used in our experiments to maintain transparency and consistency with our results.
    \item To promote transparency and facilitate further research, we commit to sharing our code on anonymous GitHub now and will open source after our paper is accepted.
\end{itemize}

\bibliography{latex/custom}

\newpage
\appendix


\section{Implementation Details}
\label{app:imple_details}

\subsection{Training Hyperparameters}
\paragraph{Supervise learning phase.} For all compared methods, we kept the same hyperparameters.
We set learning rate $\eta = 1\mathrm{e}{-}5$,  a maximum of 8 epochs, the optimizer AdamW~\citep{Loshchilov2017adamw} with $\beta_1 =0.9$ and $\beta_2=0.95$ and the cutoff length of 20000 tokens. 
Table~\ref{tab:hyperparameters} shows the detailed information on method-specific
hyperparameters for baselines.

\begin{table}[h]
\vspace{-5pt}
\centering

\begin{tabular}{lc}
\toprule
\textbf{Hyperparameter} & \textbf{DAPO} \\
\midrule
Learning Rate          & $1\times10^{-6}$ \\
Sequence Length        & 8k \\
Train Batch Size       & 256 \\
PPO Mini-batch Size    & 32 \\
Total Steps            & 200 \\
Rollout Numbers        & 16 \\
Others                 & $c_l=0.2,\; c_h=0.28$ \\
\bottomrule

\end{tabular}
\caption{Training hyperparameters for RL training.}
\label{tab:hyperparameters_dapo}
\vspace{-16pt}
\end{table}

\paragraph{Reinforcement learning phase.}
We selected the last checkpoint of all the supervised models as the cold start point for RL.
For the subsequent RL phase, we employed the same hyperparameters and the same reinforcement learning algorithm \textit{DAPO}~\citep{yu2025dapo}. The details are shown in Table~\ref{tab:hyperparameters_dapo}.



\begin{table}[h]
\centering
\caption{Training hyperparameters of different models}
\label{tab:hyperparameters}
\resizebox{0.45\textwidth}{!}{
\begin{tabular}{ll}
\toprule
\textbf{Method} & \textbf{Hyperparameters} \\
\midrule
SFT & 
\begin{tabular}[t]{l}
Learning Rate: $1\mathrm{e}{-}5$ \\
Cutoff Length: 20000 \\
Gradient Accumulation Steps: 8 \\
Batch Size: 1 \\
Epochs: 8 \\
Others: -
\end{tabular} \\
\midrule
DFT & 
\begin{tabular}[t]{l}
Learning Rate: $1\mathrm{e}{-}5$ \\
Cutoff Length: 20000 \\
Gradient Accumulation Steps: 8 \\
Batch Size: 1 \\
Epochs: 8 \\
Others: -
\end{tabular} \\
\midrule
IW-SFT & 
\begin{tabular}[t]{l}
Learning Rate: $1\mathrm{e}{-}5$ \\
Cutoff Length: 20000 \\
Gradient Accumulation Steps: 8 \\
Batch Size: 1 \\
Epochs: 8 \\
Others: -
\end{tabular} \\
\midrule
PSFT & 
\begin{tabular}[t]{l}
Learning Rate: $1\mathrm{e}{-}5$ \\
Cutoff Length: 20000 \\
Gradient Accumulation Steps: 8 \\
Batch Size: 1 \\
Epochs: 8 \\
Others: $kl=0.0, c_l=0.2, c_h=0.28$
\end{tabular} \\
\midrule
\ourtitle & 
\begin{tabular}[t]{l}
Learning Rate: $1\mathrm{e}{-}5$ \\
Cutoff Length: 20000 \\
Gradient Accumulation Steps: 8 \\
Batch Size: 1 \\
Epochs: 8 \\
Others: $\lambda_H = 0.05, \lambda_{KL}=0.05, \rho=0.2$
\end{tabular} \\
\bottomrule
\end{tabular}
}
\end{table}
\paragraph{Training set up.}
We train all the methods with \textbf{8 NVIDIA H20 SXM GPUs}. For the supervise learning phase, we set \textit{gradient accumulation steps} to 8, \textit{per
device train batch size} to 1, and \textit{bfloat16 }used.
We employ the SFT and {\ourtitle} algorithms based on LLaMA-Factory~\citep{zheng2024llamafactory}.
For DFT, PSFT and IW-SFT, we use the architectures each researcher provided~\citep{wu2025dft, zhu2025psft, qin2025iwsft}.
During the subsequent RL phase, we implement DAPO algorithms based on Verl~\citep{sheng2025hybridflow}.



\begin{table*}[t]
\centering
\small
\renewcommand{\arraystretch}{1.2}

\resizebox{\textwidth}{!}{%
\begin{tabular}{l|cc cc | cc cc}
\hline
\multirow{2}{*}{\textbf{Model}} & \multicolumn{2}{c}{\textbf{AIME}} & \multicolumn{2}{c|}{\textbf{AIME25}} & \multicolumn{2}{c}{\textbf{AMC}} & \multicolumn{2}{c}{\textbf{HMMT25}} \\
 & pass@1 & pass@32 & pass@1 & pass@32 & pass@1 & pass@32 & pass@1 & pass@32 \\
\hline
\multicolumn{9}{c}{\textbf{\textit{Qwen3-4B}}} \\ 
\hline
\ourtitle & \textbf{45.7} & \textbf{73.3} & \textbf{33.4} & \textbf{60.0} & \textbf{68.9} & \textbf{90.4} & \textbf{18.5} & \textbf{50.0} \\
\hline
SFT w/ Entropy \& KL Regularization & 41.7 & 70.0 & 30.2 & 53.3 & 67.7 & 87.9 & 17.1 & 43.3 \\
\hline
\multicolumn{9}{c}{\textbf{\textit{Qwen3-8B}}} \\ 
\hline
\ourtitle & \textbf{47.2} & \textbf{80.0} & \textbf{34.8} & \textbf{66.7} & \textbf{70.8} & \textbf{90.4} & \textbf{19.6} & \textbf{50.0} \\
\hline
SFT w/ Entropy \& KL Regularization & 44.6 & 76.7 & 32.4 & 56.7 & 67.5 & 87.9 & 17.6 & 43.3 \\
\hline
\end{tabular}
}
\caption{Ablation study of masking mechanism.}
\label{tab:ablation_masking}
\end{table*}

\subsection{Inference Hyperparameters}
During the evaluation, we adopt that the max response length is set to 8k, and \textit{temperature} is set to 1.0. 
To avoid high variance in results and ensure fair comparisons, we report avg@32, i.e., pass@1, shown in our results, on four benchmarks by sampling 32 times.

\section{Results detail}
\subsection{Pass@k analysis on Qwen3-4B}
\label{app:more_results}

\begin{figure}[htbp]
\vspace{-5pt}
    \centering
    \includegraphics[width=1.0\linewidth]{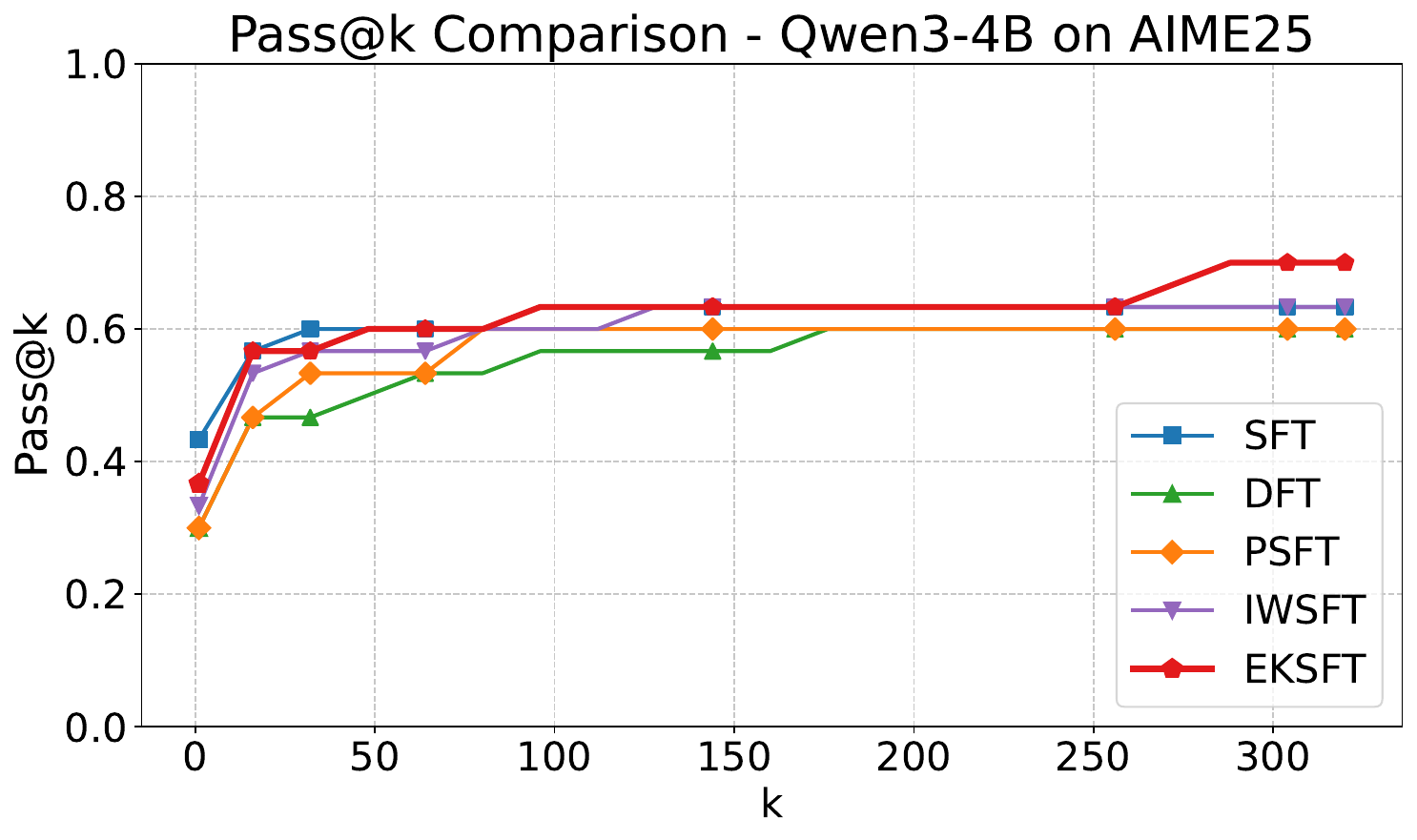}
    \caption{We compared the pass@k on AIME25 based on Qwen3-4B model.}
    \label{fig:passk_4B}
    \vspace{-16pt}
\end{figure}

As shown in Figure~\ref{fig:passk_4B}, we also compared the pass@k on AIME25 by extending the $k$ up to 320 on the Qwen3-4B. 
{\ourtitle} consistently achieved
higher pass@k performance when $k$ is greater than 250,
indicating stronger exploration and reasoning diversity.

\subsection{Ablation on the masking mechanism}
\label{app:masking_mech}

As shown in Table~\ref{tab:ablation_masking}, we compared {\ourtitle} with a global-regularized SFT baseline that applies entropy and KL regularization to all tokens.
Across both Qwen3-4B and Qwen3-8B, {\ourtitle} consistently outperforms this baseline on all benchmarks in terms of both pass@1 and pass@32.
For the Qwen3-4B, {\ourtitle} achieves \textbf{+6.7\%} improvement on AIME25 and HMMT25 pass@32,
while for the Qwen3-8B, {\ourtitle} still shows notable improvements on AIME25 pass@32 (\textbf{+10.0\%}) and HMMT25 pass@32 (\textbf{+6.7\%}).
These results suggest that the benefits of {\ourtitle} are not solely due to entropy/KL regularization,
but critically, from the selective exclusion of high-entropy or high-KL tokens during token-level imitation, which is crucial for improving both accuracy and exploration.
\begin{figure}[htbp]
    \vspace{-12pt}
    \centering
    \includegraphics[width=0.95\linewidth]{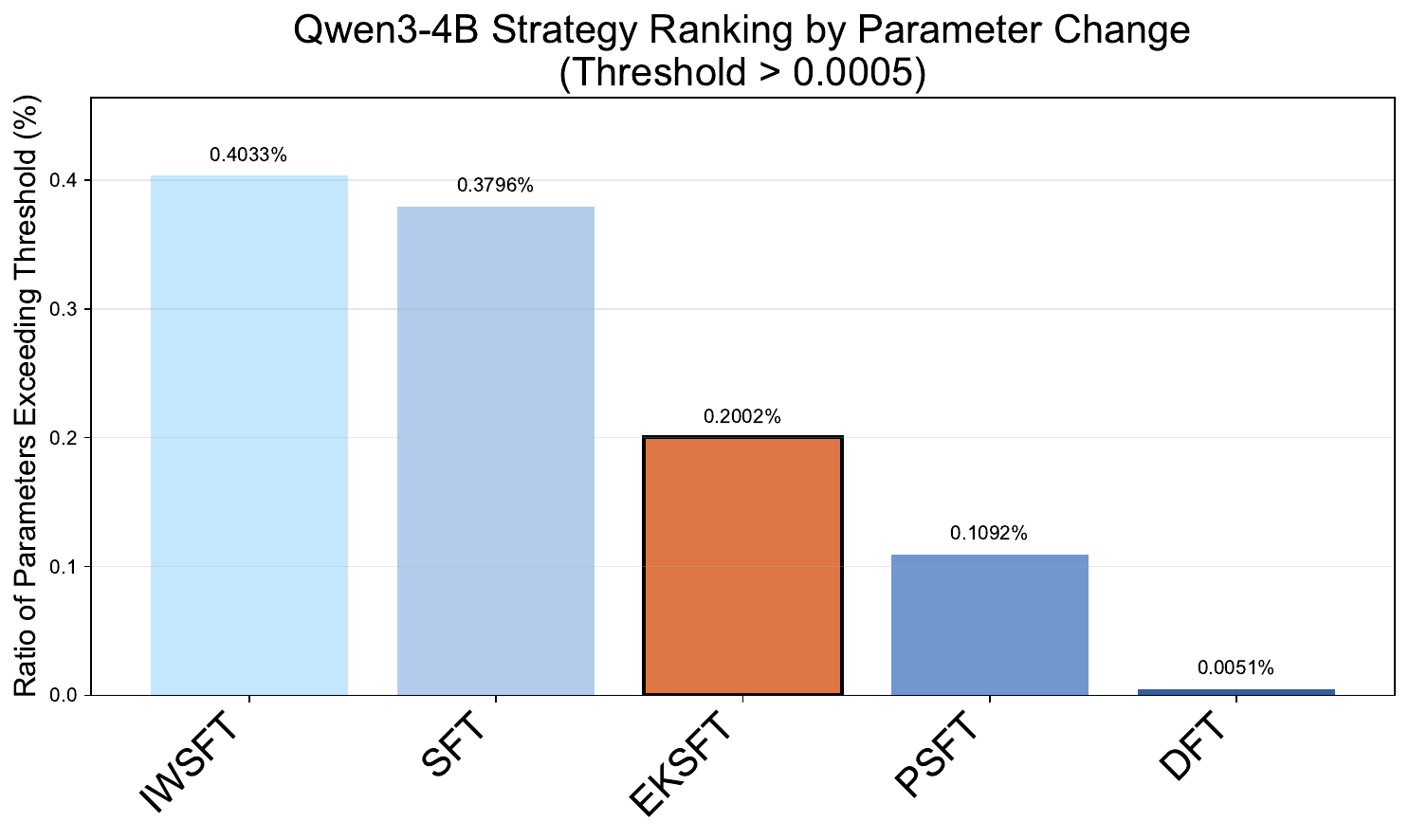}
    \caption{We reported the relative changes of the trained models' parameters, compared to the base model on Qwen3-4B.}
    \label{fig:parameter_4B}
    \vspace{-16pt}
\end{figure}

\section{Parameters Drift} \label{app:parameter_drift}
Combing Table~\ref{tab:stage1} with Figure~\ref{fig:parameter_4B}, 
we observe that standard SFT and IW-SFT introduce the largest parameter changes, and while they improve over the base model, their gains come with substantial parameters drift that may impair the pretrained distribution and exploration capacity.
In contrast, {\ourtitle} achieves the best overall performance on Qwen3-4B, 
while exhibiting much smaller parameter drift than SFT and IW-SFT.
This suggests that {\ourtitle} attains stronger exploration during the subsequent RL phase, as shown in Table~\ref{tab:stage2}.

Interestingly, PSFT and DFT yield even smaller parameter changes than {\ourtitle}, yet their performance is worse than {\ourtitle}, indicating that minimizing drift alone is insufficient,
since overly conservative updates will fail to elicit task-relevant behaviors.
Taken together these four methods, these results support the notion of \textbf{effective drift}.
Specifically, {\ourtitle} employs the masking mechanism that excludes the high entropy and high KL-divergence tokens from imitation and introducing the entropy regularization and KL regularization for the masked tokens can make task-targeted updates and lead to minimal but purposeful parameter movement.




\begin{figure}[htbp]
    \centering
    \includegraphics[width=1.0\linewidth]{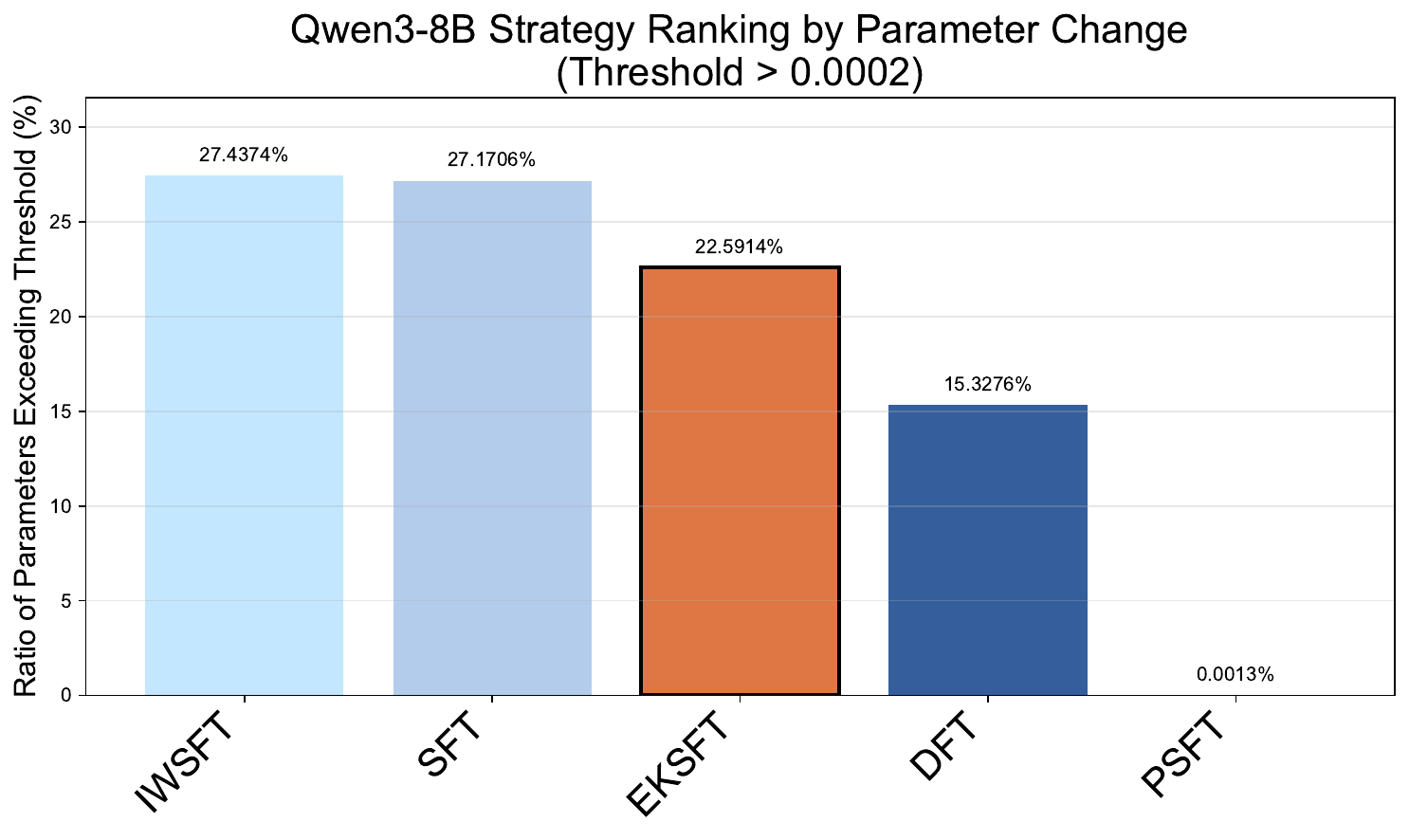}
    \caption{We reported the relative changes of the trained models' parameters, compared to the base model on Qwen3-8B.}
    \label{fig:parameter_8B}
    \vspace{-5pt}
\end{figure}


\begin{figure*}[htbp]
    \centering
    \begin{minipage}{0.99\linewidth}
    \includegraphics[width=\linewidth]{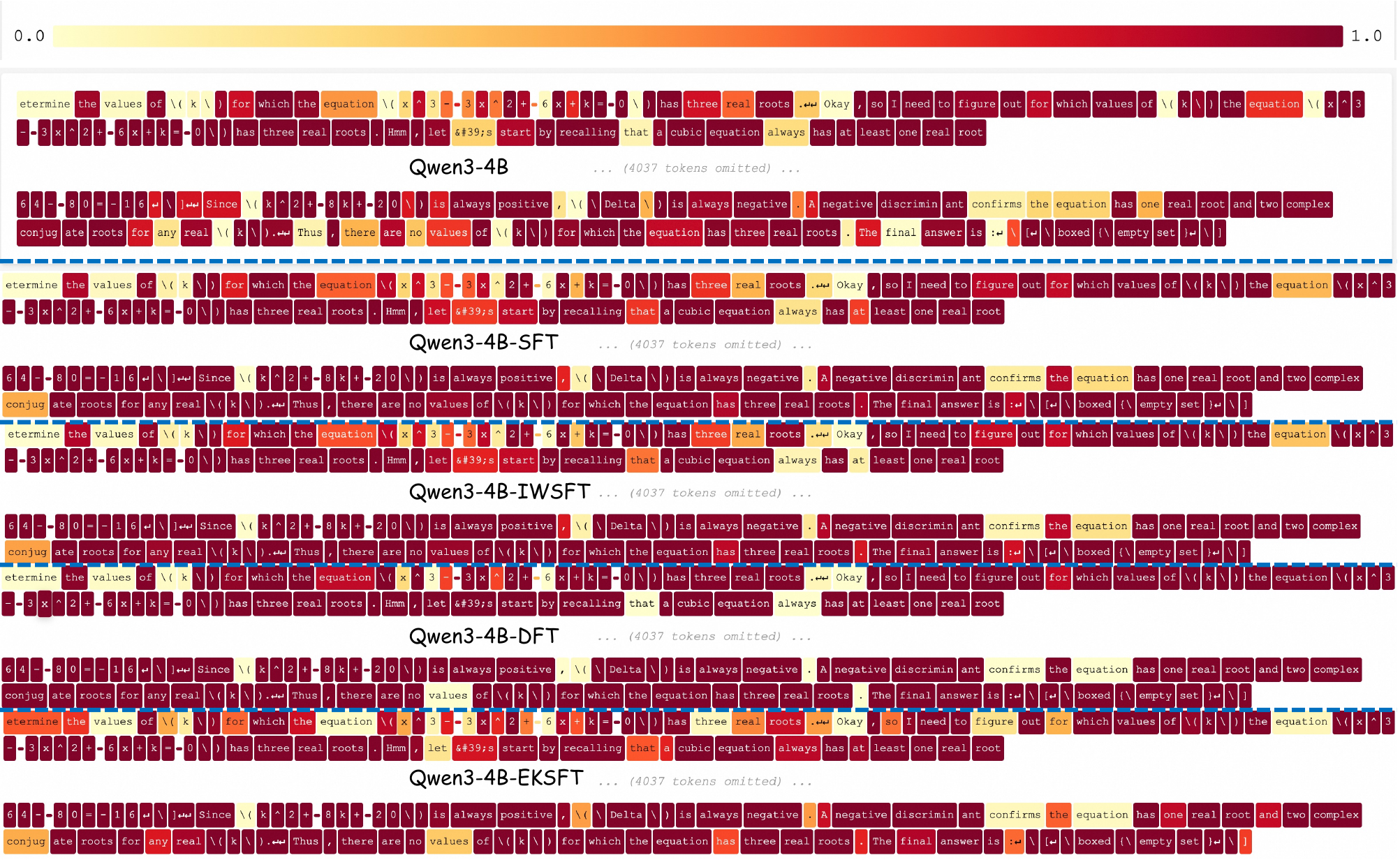}
    \caption{We reported the token probabilities of different methods on the same question with its solution based on Qwen3-4B.}
    \label{fig:case_study}
    \end{minipage}
\end{figure*}
For Qwen3-8B, Figure~\ref{fig:parameter_8B} shows a similar drift pattern. 
Standard SFT and IW-SFT produce the largest parameter changes, whereas {\ourtitle} substantially reduces the fraction of parameters exceeding the change threshold, indicating better preservation of the pretrained distribution. 
When combined with the results in Table~\ref{tab:stage1}, {\ourtitle} also achieves the strongest overall performance on Qwen3-8B, outperforming SFT and IW-SFT despite incurring less drift,
validating the effectiveness of {\ourtitle} across both Qwen3-4B and Qwen3-8B.

Overall, combining Table~\ref{tab:stage1}, Table~\ref{tab:stage2}, Figure~\ref{fig:parameter_4B} and Figure~\ref{fig:parameter_8B}, we find that {\ourtitle} consistently delivers the best or near-best performance in both the supervised learning stage and the subsequent RL stage, while inducing substantially less parameter drift than standard SFT-style training. 
Meanwhile, PSFT and DFT can reduce drift further but do not improve performance, indicating that minimizing drift alone is insufficient.
These results support our motivation that selectively excluding high-entropy and high-KL tokens 
from imitation enables effective and activates task-relevant knowledge without reducing diversity.

\section{Case Study}

In this section, we reported the token probabilities of different methods when using the same question and its ground truth.
Figure~\ref{fig:case_study} illustrates the next-token probability assigned to the ground-truth token along the reasoning trace, where darker red indicates higher probability.
Standard SFT (along with IWSFT and DFT) results in a uniform increase in saturation across the sequence, causing most tokens to become darker red,
indicating that the model indiscriminately assigns high probability to the demonstrated tokens.
This probability overconfidence reflects a sharpened distribution and reduces sampling diversity in the subsequent usage. 


In contrast, {\ourtitle} exhibits a structured probability reallocation, which excludes the high entropy and KL-divergence tokens from imitation,
avoiding the dark red saturation observed in SFT, preventing mode collapse and maintaining distribution.
This aligns with our motivation to activate the task-relevant knowledge in favor of reliable reasoning,
ultimately preserving the exploration capability essential for subsequent RL.

\section{Theoretical Analysis of Selective Masking}
\label{app:theory}
\hl{
In this section we provide a token-level formal analysis that justifies the design of {\ourtitle}: (i) why standard cross-entropy (CE) imitation is most harmful precisely on high-entropy and high-KL tokens, (ii) why replacing CE on the masked set with our entropy/KL regularization yields a label-free, low-variance update, and (iii) why selective regularization on $\mathcal{M}$ is preferable to applying the same regularization globally.
}

\subsection{High-entropy tokens dominate the SFT gradient}
\label{app:theory_grad}
\hl{
For a token position $t$, standard SFT optimizes}
\begin{equation}
\ell_t^{\mathrm{SFT}}(\theta) = -\log \pi_\theta(y_t \mid x_{<t}).
\end{equation}
\hl{Let $z_t$ denote the pre-softmax logits and $\pi_\theta = \mathrm{softmax}(z_t)$. The gradient with respect to the logits is}
\begin{equation}
\frac{\partial \ell_t^{\mathrm{SFT}}}{\partial z_t} = \pi_\theta(\cdot \mid x_{<t}) - e_{y_t},
\end{equation}
\hl{where $e_{y_t}$ is the one-hot indicator at the gold token. Therefore}
\begin{equation}
\Big\|\frac{\partial \ell_t^{\mathrm{SFT}}}{\partial z_t}\Big\|_2^2
= \sum_{v\in\mathcal{V}} \pi_\theta(v)^2 + 1 - 2\pi_\theta(y_t).
\end{equation}
\hl{Using $\sum_v \pi_\theta(v)^2 \le 1$, we obtain}
\begin{equation}
\Big\|\frac{\partial \ell_t^{\mathrm{SFT}}}{\partial z_t}\Big\|_2^2 \le 2\bigl(1 - \pi_\theta(y_t)\bigr).
\end{equation}
\hl{At a high-entropy position, $\pi_\theta$ is close to uniform, so $\pi_\theta(y_t)$ is small and the residual $\|\pi_\theta - e_{y_t}\|$ stays at an $O(1)$ scale. In the uniform limit $\pi_\theta(v)\approx 1/|\mathcal{V}|$,}
\begin{equation}
\Big\|\frac{\partial \ell_t^{\mathrm{SFT}}}{\partial z_t}\Big\|_2^2 \approx 1 - \frac{1}{|\mathcal{V}|}.
\end{equation}
\hl{By contrast, at a low-entropy, high-confidence position with $\pi_\theta(y_t) = 1-\varepsilon$,}
\begin{equation}
\|\pi_\theta - e_{y_t}\|_2^2 \approx 2\varepsilon^2,
\end{equation}
\hl{which is much smaller. High-entropy tokens therefore generate disproportionately large CE gradients and tend to dominate parameter updates in the low-data SFT regime, accelerating distribution sharpening and parameter drift. A symmetric argument applies to high-KL tokens: such positions are exactly where $\pi_\theta$ already deviates from $\pi_{\mathrm{ref}}$, so further label-driven imitation pushes the policy further away from the pretrained distribution. Masking these tokens removes the high-energy, unstable component of the gradient and reallocates the limited update budget to more stable, transferable positions.}

\subsection{Label-free regularization on the masked set}
\label{app:theory_label_free}

\hl{On the masked set $\mathcal{M}$, instead of $-\log\pi_\theta(y_t)$ we optimize}
\begin{equation}
\mathcal{L}^{\mathrm{reg}}_{\mathcal{M}}(\theta) = \sum_{t\in\mathcal{M}}\Bigl[\lambda_{\mathrm{KL}}\, \mathrm{KL}(\pi_\theta\,\|\,\pi_{\mathrm{ref}}) - \lambda_H\, H(\pi_\theta)\Bigr].
\end{equation}
\hl{Expanding $\mathrm{KL}(\pi_\theta\|\pi_{\mathrm{ref}}) = -H(\pi_\theta) + H(\pi_\theta, \pi_{\mathrm{ref}})$ with $H(\pi_\theta,\pi_{\mathrm{ref}}) = -\sum_v \pi_\theta(v)\log\pi_{\mathrm{ref}}(v)$, we obtain}
\begin{equation}
\begin{split}
\mathcal{L}^{\mathrm{reg}}_{\mathcal{M}}(\theta) 
&= \sum_{t\in\mathcal{M}}\Bigl[-(\lambda_{\mathrm{KL}}+\lambda_H)\, H(\pi_\theta) \\
&\qquad\quad + \lambda_{\mathrm{KL}}\, H(\pi_\theta, \pi_{\mathrm{ref}})\Bigr].
\end{split}
\end{equation}
\hl{Letting $J_v = \nabla\log\pi_\theta(v)$, the standard SFT gradient at token $t$ can be written as}
\begin{equation}
\nabla \ell_t^{\mathrm{SFT}} = \sum_{v\in\mathcal{V}} \pi_\theta(v)\, J_v - J_{y_t},
\end{equation}
\hl{which contains the explicit one-hot forcing term $-J_{y_t}$. In contrast, the gradient of $\mathcal{L}^{\mathrm{reg}}_{\mathcal{M}}$ at any masked token has the form}
\begin{equation}
\nabla R_t = \sum_{v\in\mathcal{V}} a_v(\theta)\, J_v,
\end{equation}
\hl{where the coefficients $a_v(\theta)$ depend only on $\pi_\theta$, $\log\pi_\theta$, and $\log\pi_{\mathrm{ref}}$. Crucially, no term of the form $-J_{y_t}$ appears: the update is \emph{label-free}. Hence on the most uncertain or already-drifted positions our update no longer contains the high-variance, ``exogenous forcing'' direction that drives entropy collapse and parameter drift. Instead, it balances two soft forces---staying close to $\pi_{\mathrm{ref}}$ and maintaining higher entropy---scaled linearly by $\lambda_{\mathrm{KL}}$ and $\lambda_H$. This is consistent with the smaller drift we observe empirically (Appendix}~\ref{app:parameter_drift}).

\subsection{Why selective regularization beats global regularization}
\label{app:theory_selective}
\hl{
A natural alternative is to apply entropy/KL regularization \emph{globally}, on every token, alongside full CE supervision. We argue this is suboptimal in our setting and our experiments confirm it (Appendix}~\ref{app:masking_mech}).
\hl{
Decompose the per-token update under global regularization into two parts: a CE part proportional to $\pi_\theta - e_{y_t}$ and a regularization part proportional to $\nabla R_t$. On low-entropy, low-KL tokens, the CE part is small (Appendix}~\ref{app:theory_grad}) \hl{and reliably injects task-relevant signal; adding regularization here mostly \emph{cancels} useful gradient and makes updates overly conservative, particularly damaging in the low-data regime. On high-entropy, high-KL tokens, both parts are large and partially counteract each other, leaving training dynamics dominated by the high-variance CE component.

Selective masking instead concentrates the constraint budget where it is most needed. On the masked set $\mathcal{M}$, the CE term is \emph{removed entirely} (not merely offset), so the strong forcing direction $-J_{y_t}$ is eliminated and only the soft, label-free regularizer remains. On the complement $\overline{\mathcal{M}}$, the clean CE signal is left intact, free of regularization-induced cancellation. This split---hard control where risk is high, undisturbed learning where signal is clean---is what enables {\ourtitle} to simultaneously activate task-relevant knowledge and preserve exploration capacity.}

\section{Complementarity of Entropy and KL Token Sets}
\label{app:iou}
\hl{
A natural concern is whether the high-entropy and high-KL token sets are largely the same, in which case using both signals would be redundant. To test this, during training we logged per-token entropy and per-token KL for every step, formed the two top-$\rho$ sets $\mathcal{M}_H$ and $\mathcal{M}_{\mathrm{KL}}$ (with $\rho=0.2$), and computed the Intersection-over-Union $\mathrm{IoU} = |\mathcal{M}_H \cap \mathcal{M}_{\mathrm{KL}}| / |\mathcal{M}_H \cup \mathcal{M}_{\mathrm{KL}}|$ at each step.}

\begin{table}[h]
\centering
\small
\begin{tabular}{lc}
\toprule
\textbf{Statistic} & \textbf{IoU} \\
\midrule
Max & 0.59 \\
Min & 0.09 \\
Avg & 0.50 \\
\bottomrule
\end{tabular}
\caption{\hl{IoU between the high-entropy token set $\mathcal{M}_H$ and the high-KL token set $\mathcal{M}_{\mathrm{KL}}$ over training steps.}}
\label{tab:iou}
\end{table}
\hl{
With an average IoU of about $0.50$, only roughly half of the tokens overlap between the two sets. In other words, positions where the model is uncertain (high entropy) are not necessarily the positions that deviate most from the reference distribution (high KL). The two signals capture different aspects of training dynamics, which is why combining them via union (Equation}~\ref{eq.11}) \hl{yields strictly more coverage than either single signal alone.}

\section{Sensitivity to Masking Ratio $\rho$}
\label{app:ratio}
\hl{
We sweep the masking ratio $\rho \in \{0.0, 0.1, 0.2, 0.3, 0.4\}$ on Qwen3-4B and report pass@1 and pass@32 on AIME, AIME25, and AMC.}

\begin{table}[h]
\centering
\small
\resizebox{0.48\textwidth}{!}{%
\begin{tabular}{c|cc|cc|cc}
\toprule
\multirow{2}{*}{$\rho$} & \multicolumn{2}{c|}{\textbf{AIME}} & \multicolumn{2}{c|}{\textbf{AIME25}} & \multicolumn{2}{c}{\textbf{AMC}} \\
 & pass@1 & pass@32 & pass@1 & pass@32 & pass@1 & pass@32 \\
\midrule
0.0 & 46.3 & 73.3 & 32.1 & 56.7 & 68.7 & 86.7 \\
0.1 & 46.0 & 73.3 & 32.3 & 56.7 & 68.7 & 86.7 \\
0.2 & \textbf{45.7} & \textbf{73.3} & \textbf{33.4} & \textbf{60.0} & \textbf{68.9} & \textbf{90.4} \\
0.3 & 45.2 & 70.0 & 32.5 & 60.0 & 65.6 & 85.5 \\
0.4 & 35.0 & 60.0 & 24.0 & 46.7 & 58.7 & 80.1 \\
\bottomrule
\end{tabular}}
\caption{\hl{Sensitivity of {\ourtitle} to the masking ratio $\rho$ on Qwen3-4B. Best per column in bold (excluding the $\rho{=}0.0$ row, which corresponds to no masking).}}
\label{tab:ratio}
\end{table}
\hl{
The results show a clear sweet spot around $\rho = 0.2$: small $\rho$ behaves close to standard SFT and provides little benefit, while $\rho = 0.4$ collapses performance across all three benchmarks (e.g., a $-11.3\%$ pass@1 drop on AIME relative to $\rho=0.0$). This is expected, since excessive masking removes too much supervision signal and prevents the model from acquiring task-relevant patterns. A moderate ratio (around $0.2$) strikes the right balance between preserving pretrained capabilities and providing adequate supervision for capability activation, which is why we adopt $\rho = 0.2$ throughout the paper.}

\section{Comparison with Random Masking}
\label{app:random_mask}
\hl{
To verify that the gains of {\ourtitle} come from \emph{which} tokens are masked rather than from masking per se, we compare {\ourtitle} with a Random Masking baseline that uniformly drops $10\%$ of tokens from CE supervision and applies the same entropy/KL regularization on the dropped tokens. All other settings are kept identical.}

\begin{table}[h]
\centering
\small
\resizebox{0.48\textwidth}{!}{%
\begin{tabular}{l|cc|cc|cc}
\toprule
\multirow{2}{*}{\textbf{Method}} & \multicolumn{2}{c|}{\textbf{AIME}} & \multicolumn{2}{c|}{\textbf{AIME25}} & \multicolumn{2}{c}{\textbf{AMC}} \\
 & pass@1 & pass@32 & pass@1 & pass@32 & pass@1 & pass@32 \\
\midrule
{\ourtitle} (Ours) & \textbf{45.7} & \textbf{73.3} & \textbf{33.4} & \textbf{60.0} & \textbf{68.9} & \textbf{90.4} \\
Random Masking (10\%) & 35.0 & 60.0 & 26.5 & 56.7 & 66.0 & 85.5 \\
\bottomrule
\end{tabular}}
\caption{\hl{Comparison with a Random Masking baseline on Qwen3-4B.}}
\label{tab:random_mask}
\end{table}
\hl{
Random Masking lags {\ourtitle} on every benchmark and metric, with particularly large gaps on AIME pass@1 ($-10.7\%$) and AIME pass@32 ($-13.3\%$). Indiscriminate masking discards task-relevant evidence and introduces unstructured noise, while {\ourtitle} concentrates masking on positions that are demonstrably risky for entropy collapse and parameter drift. This confirms that the benefit of {\ourtitle} stems from the entropy/KL-guided selection criterion, not merely from reducing the number of supervised tokens.}

\section{Generalization to Tool-Use Benchmarks}
\label{app:bfcl}
\hl{To assess whether the benefits of {\ourtitle} generalize beyond mathematical reasoning, we additionally evaluate it on the tool-use setting. We train on AceBench}~\citep{chen2025acebench} \hl{under the same training configuration as the main experiments, and evaluate on the widely used BFCL}~\citep{patil2025bfcl} \hl{\textit{live} and \textit{Irrelevance Detection} subsets, which respectively measure overall tool-use performance and robustness to irrelevant requests.}

\begin{table}[h]
\centering
\small
\begin{tabular}{l|cc}
\toprule
\textbf{Method (Qwen3-4B)} & \textbf{BFCL live} & \textbf{Irrelevance Det.} \\
\midrule
Base & 82.4 & 75.9 \\
SFT & 84.4 & 89.7 \\
DFT & 70.5 & 93.5 \\
{\ourtitle} (Ours) & \textbf{86.2} & \textbf{100.0} \\
\bottomrule
\end{tabular}
\caption{\hl{BFCL results on Qwen3-4B trained on AceBench.}}
\label{tab:bfcl_4b}
\end{table}

\begin{table}[h]
\centering
\small
\begin{tabular}{l|cc}
\toprule
\textbf{Method (Qwen3-8B)} & \textbf{BFCL live} & \textbf{Irrelevance Det.} \\
\midrule
Base & 84.5 & 76.9 \\
SFT & 86.6 & 99.5 \\
DFT & 82.5 & \textbf{100.0} \\
{\ourtitle} (Ours) & \textbf{88.0} & \textbf{100.0} \\
\bottomrule
\end{tabular}
\caption{\hl{BFCL results on Qwen3-8B trained on AceBench.}}
\label{tab:bfcl_8b}
\end{table}
\hl{
Across both model sizes, {\ourtitle} achieves the best BFCL live score and saturates Irrelevance Detection at $100.0$. Notably, DFT improves Irrelevance Detection but degrades on BFCL live (especially on Qwen3-4B, $-13.9\%$ versus base), indicating that aggressive reweighting can harm in-task tool-use accuracy. {\ourtitle} avoids this trade-off and improves both metrics simultaneously, demonstrating that its benefits transfer to unseen tasks and datasets beyond mathematical reasoning.}

\end{document}